\documentclass[letterpaper, 10pt, conference]{ieeeconf}  

\IEEEoverridecommandlockouts                              

\usepackage{graphics} 
\usepackage{epsfig} 
\usepackage{times} 
\usepackage{amsmath} 
\usepackage{amssymb}  
\usepackage[whole]{bxcjkjatype}
\usepackage{bm}
\usepackage{color}
\usepackage{mathabx}
\usepackage{multirow}

\newcommand*{\cb}[1]{{\color{blue} \textbf{#1}}}
\newcommand*{\re}[1]{{\color{red} \textbf{#1}}}

\title{\LARGE \bf
Data-Driven Stochastic Motion Evaluation and Optimization with Image\\
by Spatially-Aligned Temporal Encoding
}

\author{Takeru Oba$^{1}$ and Norimichi Ukita$^{1}$
\thanks{*This work was not supported by any organization.}
\thanks{$^{1}$The authors are with Graduate School of Engineering, Toyota Technological Institute, 2-12-1 Hisakata, Tempaku, Nagoya, 468-8511 Japan,
        {\tt\small sd21502@toyota-ti.ac.jp}}%
}

\begin{document}

\maketitle
\thispagestyle{empty}
\pagestyle{empty}


\begin{abstract}
This paper proposes a probabilistic motion prediction method for long motions.
The motion is predicted so that it accomplishes a task from the initial state observed in the given image.
While our method evaluates the task achievability by the Energy-Based Model (EBM), previous EBMs are not designed for evaluating the consistency between different domains (i.e., image and motion in our method).
Our method seamlessly integrates the image and motion data into the image feature domain by spatially-aligned temporal encoding so that features are extracted along the motion trajectory projected onto the image.
Furthermore, this paper also proposes a data-driven motion optimization method, Deep Motion Optimizer (DMO), that works with EBM for motion prediction.
Different from previous gradient-based optimizers, our self-supervised DMO alleviates the difficulty of hyper-parameter tuning to avoid local minima.
The effectiveness of the proposed method is demonstrated with a variety of experiments with similar SOTA methods.
\end{abstract}

\section{INTRODUCTION}

%
One of the difficulties in motion prediction is that, for a certain situation, there might be a variety of appropriate motions, each of which accomplishes the task.
It is not easy to train such one-to-many relationships compared to one-to-one relationships.
In particular, a deterministic model is not good at training the one-to-many relationships.
For representing the one-to-many relationships for motion prediction, probabilistic generative models are commonly used~\cite{hassan2021stopred, cao2020long}.

\begin{figure}[t]
  \begin{center}
    \includegraphics[width=1.0\columnwidth]{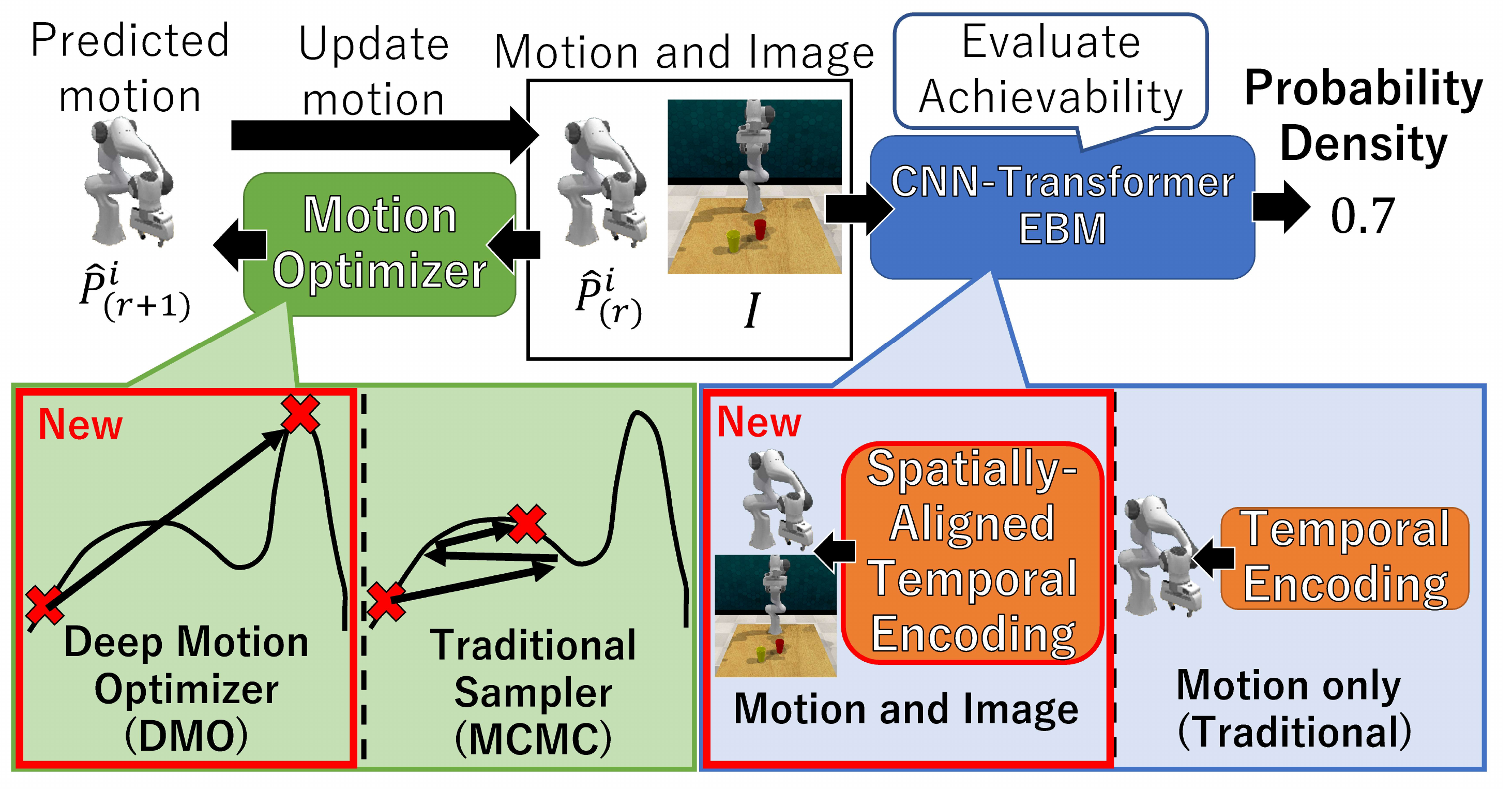}
    \caption{Overview of our proposed method. CNN-Transformer EBM estimates the probability of a motion. The motion is efficiently optimized to improve its probability by Deep Motion Optimizer.}
    \label{fig:proposed method overall}
  \end{center}
\end{figure}

An Energy-Based Model (EBM) is governed by an energy function representing the probability density of a state.
EBM is more flexible than other generative models such as VAE~\cite{kingma2014vae,vahdat2020nvae} and Normalizing Flow (NF)~\cite{dinh2016realnvp, papamakarios2017masked} in terms of the model architecture and the prior distribution.
%
As shown in the right part of Fig.~\ref{fig:proposed method overall},
given an image observing the initial situation of a robot and its surrounding environment, EBM estimates the probability density of consistency between the initial image and motion.
The probability density becomes higher, if the motion is considered to be appropriate for accomplishing the task from the initial situation.

However, different from other generative models, EBM predicts no motion but estimates only the probability density of each motion.
With the probability density estimated by EBM, appropriate motions can be predicted using sampling such as MCMC.
Accordingly, the accuracy of the predicted motion depends on EBM as well as sampling.
However, many sampling methods are controlled by hyper-parameters, and it is difficult to manually tune them to avoid local minima in high-dimensional distributions such as long sequences.

To resolve the aforementioned problem of parameter tuning, our proposed Deep Motion Optimizer (DMO) rectifies each motion with a data-driven manner using a set of training motions.
DMO is trained to directly rectify the motion to its ideal one, while sampling-based methods iteratively search for the ideal one based on the gradients of the probability density distribution, as illustrated in the left of Fig.~\ref{fig:proposed method overall}.

Another difference between our EBM and previous EBMs for motion prediction is the types of domains fed into EBM.
Since previous EBMs~\cite{pang2021trajectory} evaluate the probability density of consistency between the same domain data (i.e., the observed and predicted motions), it is relatively easy to evaluate the consistency.
In our EBM, on the other hand, the consistency is evaluated between data of different domains (i.e., the initial image and the predicted motion).
While the image is useful for motion prediction based on the surrounding environment, we have to properly evaluate the consistency between the image and each predicted motion.
To facilitate this evaluation in our method, the image and the predicted motion are seamlessly integrated into the image domain so that image features are extracted along the motion trajectory for spatially-aligned temporal encoding.

Our novel contributions are summarized as follows:
\begin{itemize}
\item Different from previous EBMs evaluating only motions, our proposed EBM properly evaluates the consistency between the image and motion domains by efficient spatially-aligned temporal encoding.
\item Instead of hand-crafted hyper-parameters for tuning iterative optimizers, the data-driven framework for training DMO is proposed to directly optimize the motion.
\end{itemize}


\section{Related work}
\subsection{EBM}
\label{section: Related work EBM}

It is not easy to represent a complex distribution in a high-dimensional space by EBM using sampling such as MCMC.
%
To tackle this problem, in Score matching~\cite{hyvarinen2005estimation, song2020improved} where no sampling is needed, given the gradient of the probability density of a real sample~\cite{hyvarinen2005estimation}, EBM is trained so that the gradient estimated by EBM gets close to the given real gradient.
In Noise Contrastive Estimation~\cite{gutmann2010noise, bose2018adversarial}, given a simple distribution where sampling is easy, EBM is trained to discriminate between this simple distribution and the distribution of real samples.
%
Even with these training methods~\cite{hyvarinen2005estimation, song2020improved, gutmann2010noise, bose2018adversarial}, however, EBM inference still need to estimate the complex high-dimensional distribution by sampling.

Another solution 
is to reduce the sampling difficulty both in the training and inference stages by using other generative models such as VAE~\cite{pang2021trajectory, xiao2021vaebm} and NF~\cite{gao2020flow, xie2021tale}.
In~\cite{pang2021trajectory} and~\cite{xiao2021vaebm}, sampling is achieved in the low-dimensional space induced by VAE.
In~\cite{xie2021tale}, initial inaccurate samples are produced fast by NF, and then these initial samples are given to EBM for more accurate sampling.
Even with these methods, however, accurate sampling is still difficult to predict the accurate high-dimensional long-term motions.

\subsection{Motion Prediction}
\label{section: Related work Motion Prediction}


Difficulties in our motion prediction problem include (i) long-term error accumulation, (ii) stochasticity of motions, and (iii) multi-domain data processing.

Even with recent
LSTM~\cite{hochreiter1997long, rahmatizadeh2018vision, pfeiffer2018data, wang2019imitation} and GRU~\cite{cho2014learning, martinez2017human}, it is difficult to predict a long-time complex motion due to gradient vanishing.
This problem is alleviated by Transformer~\cite{aksan2021spatio, giuliari2021transformer, kim2021transformer} using an attention mechanism~\cite{vaswani2017attention}, which keeps the gradient stable~\cite{kerg2020untangling}.

Stochastic motions can be predicted by probabilistic generative models such as VAE~\cite{aliakbarian2021contextually, wang2017robust, cao2020long} and GAN~\cite{barsoum2018hp, ho2016generative, hernandez2019human}.
However, these models do not explicitly estimate the probability density of a  motion.
This disadvantage prevents us from
optimizing the motion using maximum likelihood estimation using the probability density as the likelihood.
This problem is resolved by enabling probabilistic
models to explicitly estimate the probability density, for example, NF~\cite{scholler2021flomo, mao2021generating} and EBM~\cite{pang2021trajectory, florence2022implicit}.
However, NF and EBM still have several problems as described in Sec.~\ref{section: Related work EBM}.

For multi-domain data processing such as our method using image and motion data,
one of the simplest ways is to concatenate these data.
The expressiveness of the concatenated data can be improved by feature extraction such as convolution and pooling for the image data~\cite{zhang2018deep, codevilla2018end, codevilla2019exploring}.
However, spatial pooling makes is difficult to precisely localize the motion in the image~\cite{liu2018intriguing}.
Inconsistency between the coordinate systems of the
image and motion data is also problematic.
For the former problem,
we propose efficient high-dimensional feature extraction that focuses on the motion trajectory on the image.
Since the latter problem
can be resolved by coordinate rearrangement (e.g., motion heatmapping in the image~\cite{cao2020long}), our method also employs it.


\section{Inference and Training Methods}

\subsection{Notations}

An image and a motion sequence are denoted by $I$ and $P$, respectively.
$P$ is a set of $(T+1)$ temporal pose vectors $\bm{p}_{t}$ where $t \in \{ 0, \cdots, T \}$.
In our experiments, $\bm{p}_{t}$ is a 11-D vector consisting of the 3D location of the robot hand ($\bm{x}_{t}$), its orientation ($\bm{r}_{t}$), its status ($s_{t}$), and a time stamp $t$.
$\bm{x}_{t}$, $\bm{r}_{t}$, and $s_{t}$ are a 3D positional vector in the camera coordinate system, a 6D vector representing the 3D orientation~\cite{Zhou2019rotation}, and a scalar value between 0 and 1 representing the open (1) and close (0) status of the hand gripper, respectively.
$I^{i}$ and $P^{i}$ denote the $i$-th data, where $i \in \{1, \cdots, N \}$, in our training dataset.
In each of all pairs in the training dataset, $P^{i}$ can accomplish the task from $I^{i}$.


\subsection{Motion Prediction using EBM and DMO}
\label{section: inference}

\begin{figure}[t]
  \begin{center}
    \includegraphics[width=1.0\columnwidth]{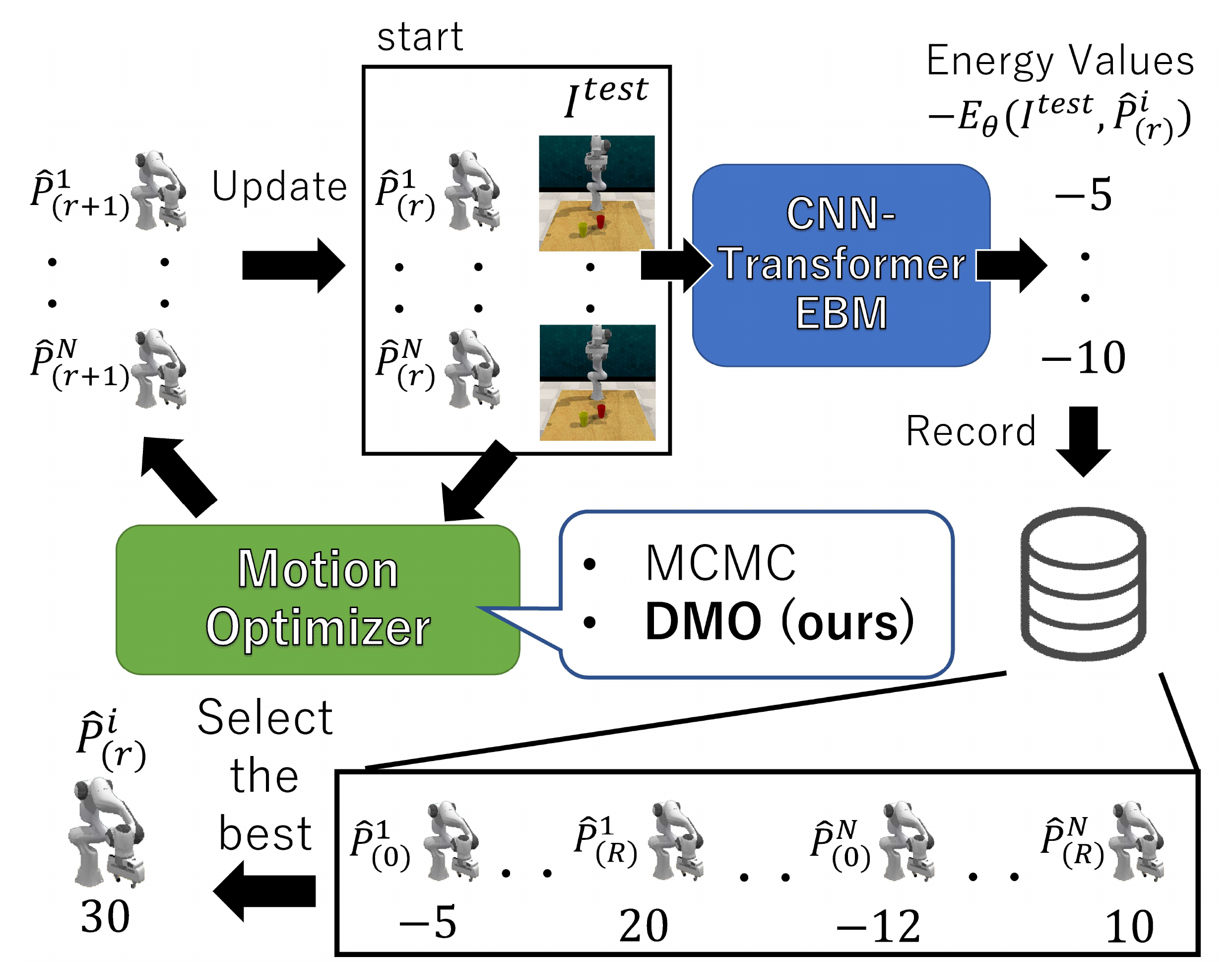}
    \caption{Motion prediction in the inference stage. 
    The image and motions are fed into EBM and the Motion Optimizer. 
    EBM estimates the energy of each motion. 
    Motion Optimizer updates the motion to increase its energy. 
    Finally, the motion having the highest energy is selected as the predicted motion.
    }
    \label{fig:inference}
  \end{center}
\end{figure}

Motion prediction from the initial state represented by the image is illustrated in Fig.~\ref{fig:inference}.
In our method, $I$ and $P$ are fed into EBM (denoted by $E_{\theta}$) to estimate the probability density $p(P|I)$ of the consistency between $I$ and $P$:
\begin{equation}
    p(P|I) = \frac{\exp(-E_{\theta}(I, P))}{Z(I)},
    \label{eq:pd}
\end{equation}
where $\theta$ denotes the parameters in EBM.
$Z$, which depends only on $I$, is defined as follows for normalization:
\begin{equation}
    Z(I) = \int \exp(-E_{\theta}(I, P)) dP
\end{equation}

In the inference stage, given $I^{test}$ and any motion $P$, 
EBM is required to estimate the consistency between $I^{test}$ and $P$.
While $Z(I)$ in Eq.~(\ref{eq:pd}) is computationally intractable, $Z(I)$ is not changed because $I$ is fixed to $I^{test}$.
Instead of the probability density, therefore, the energy term $-E_{\theta}(I^{test},P)$ is used as the output of EBM.

To increase $-E_{\theta}(I^{test},P)$, DMO updates $P$ in order to bring $P$ to its ideal motion (denoted by $P^{test}$ paired with $I^{test}$), which is not available in the inference stage, as follows:
\begin{equation}
    \hat{P}_{(1)} = DMO(I,P),
    \label{eq:dmo_1}
\end{equation}
where $I^{test}$ is substituted to $I$ in the inference stage.
Since it is not guaranteed that $\hat{P}_{(1)}$ is sufficiently close to $P^{test}$, DMO is recurrently used by rewriting Eq.~(\ref{eq:dmo_1}) to Eq.~(\ref{eq:dmo_r}): 
\begin{equation}
    \hat{P}_{(r+1)} = DMO(I,\hat{P}_{(r)}),
    \label{eq:dmo_r}
\end{equation}
where the initial motion $P$ in Eq.~(\ref{eq:dmo_1}) is denoted by $\hat{P}_{0}$, and $r$ is an integer greater than or equal to zero.

As initial motions each of which is fed into DMO in Eq.~(\ref{eq:dmo_r}), $\{ P^{1}, \cdots, P^{N} \}$ in the training dataset are reused in our method so that $\hat{P}^{i}_{(r+1)} = DMO(I^{test},\hat{P}^{i}_{(r)})$.
Among all $\{ P^{1}, \cdots, P^{N} \}$, several of them are close to $P^{test}$, while the remaining ones are far from $P^{test}$.
For efficiency, only the former motions are used.
These motions are determined by selecting the top $n$ energy values computed by EBM, and fed into DMO expressed by Eq.~(\ref{eq:dmo_r}).

In our method, both EBM and DMO are implemented as neural networks trained as described in Secs.~\ref{section: train EBM} and \ref{section: train DMO}, respectively.


\subsection{Training of EBM}
\label{section: train EBM}

If previous EBM training methods~\cite{florence2022implicit} are used for estimating $E_{\theta}$, $\theta$ is optimized by maximizing the log-likelihood of the following loss function $\mathcal{L}_{EBM}(\theta)$:
\begin{equation}
    \mathcal{L}_{EBM}(\theta) = \frac{1}{N} \sum_{i=1}^{N} \left( -E_{\theta}(I^i,P^i) - \log Z_{\theta}(I^i) \right)
\end{equation}
The gradient of $\mathcal{L}_{EBM}(\theta)$ is expressed as follows:
\begin{equation}
\begin{split}
    \nabla \mathcal{L}_{EBM}(\theta) = &\frac{1}{N} \sum_{i=1}^{N} \left( - \nabla E_{\theta}(I^i,P^i) \right. \\ 
    &\left.+ {\mathbb{E}}_{\widecheck{P}^{i} \sim p(P|I^i)} \nabla E_{\theta}(I^i, \widecheck{P}^{i}) \right), \label{equation: EBM Loss Grad}
\end{split}
\end{equation}
where $\widecheck{P}^{i}$ is a motion synthesized by sampling based on $p(P | I^{i})$ given by EBM.
This sampling process is not easy in high dimension, as described in Sec.~\ref{section: Related work EBM}.


\begin{figure}[t]
  \begin{center}
    \includegraphics[width=1.0\columnwidth]{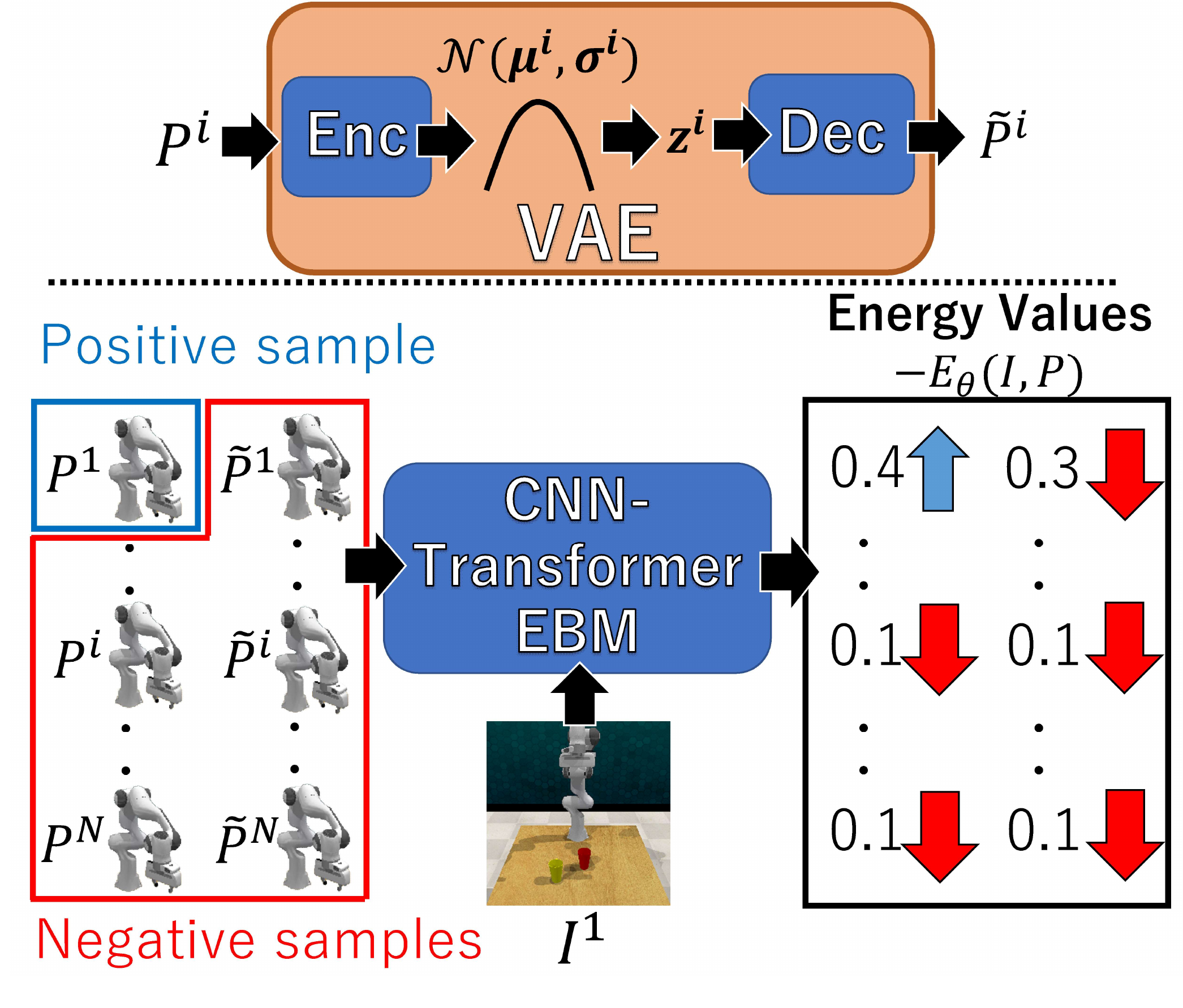}
    \vspace*{-8mm}
    \caption{Training procedure of our EBM. VAE reconstructs the motions to acquire various negative motions. 
    The image and motions are fed into the EBM. EBM estimates the energy of each motion. 
    This EBM is trained to improve the energy of the positive sample and decrease that of each negative sample.}
    \label{fig:train EBM}
  \end{center}
\end{figure}

To consider how to avoid the difficulty in this sampling, we focus on Eq.~(\ref{equation: EBM Loss Grad}).
With $- \nabla E_{\theta}(I^i, P^i)$ in Eq.~(\ref{equation: EBM Loss Grad}), $p(P^{i} | I^{i})$ as the probability density of a positive sample where $P^{i}$ accomplishes the task from $I^{i}$ is increased.
On the other hand, $\nabla E_{\theta}(I^{i}, \widecheck{P}^{i})$ decreases $p(\widecheck{P}^{i} | I^{i})$ as the probability density of a negative sample where $\widecheck{P}^{i}$ cannot achieve the task from $I^{i}$.
In our method, these positive and negative samples are selected from the training dataset as follows.
As described in Sec.~\ref{section: inference}, $\{ P^{1}, \cdots, P^{N} \}$ in the training dataset are reused as initial motions fed into DMO in the inference stage.
To make this inference succeed, $P^{i}$ is expected to be updated towards $P^{test}$, if $I^{i}$ is the closest to $I^{test}$.
On the other hand, $P^{j}$ where $j \neq i$ is expected not to be updated to $P^{i}$.
These relationships are represented in EBM so that $(I^{i}, P^{i})$ and $(I^{i}, P^{j})$ are trained as the positive and negative samples, respectively.


In addition to the positive and negative samples directly borrowed from the training dataset, our method augments samples for training EBM so that $P$ close to $P^{j}$ should be also a negative sample.
This data augmentation is done by simple low-dimensional sampling via VAE.
VAE trains an embedding space from a set of all $N$ motion sequences (i.e., $P^{1}, \cdots, P^{N}$) in the training dataset.
The trained VAE is denoted by $V$.
$P^{j}$ is fed into $V$ to get $\mathcal{N}(\bm{\mu}^{j}, \bm{\sigma}^{j})$, as shown in the upper part of Fig.~\ref{fig:train EBM}.
Let $\bm{z}^{j}$ be sampled from $\mathcal{N}(\bm{\mu}^{j}, \bm{\sigma}^{j})$.
The motion decoded from $\bm{z}^{j}$ (denoted by $\tilde{P}^{j}$) is close to its original motion $P^{j}$
Since it is easy to represent this simple Gaussian distribution in a low-dimensional embedding space, sampling from this distribution in $V$ allows us to stably synthesize $\tilde{P}^{j}$ close to $P^{j}$.

As shown in the lower part of Fig.~\ref{fig:train EBM}, with the positive and negative samples prepared by the aforementioned manner, EBM is trained with the following loss function:
\begin{equation}
\begin{split}
\nabla \mathcal{L}_{EBM}(\theta) = &\frac{1}{N} \sum_{i=1}^{N} \left( - \nabla E_{\theta}(I^i,P^i) + \sum_{j \neq i} \nabla E_{\theta}(I^i,P^j) \right. \\  
    &\left. + \sum_{j=1}^{N} {\mathbb{E}}_{\tilde{P}^{j} \sim \mathcal{N}(\bm{\mu}^{j}, \bm{\sigma}^{j})} \nabla E_{\theta}(I^{i}, \tilde{P}^{j}) \right)
\end{split}
\label{eq:EBM_loss}
\end{equation}


\subsection{Training of DMO}
\label{section: train DMO}

\begin{figure}[t]
  \begin{center}
    \includegraphics[width=1.0\columnwidth]{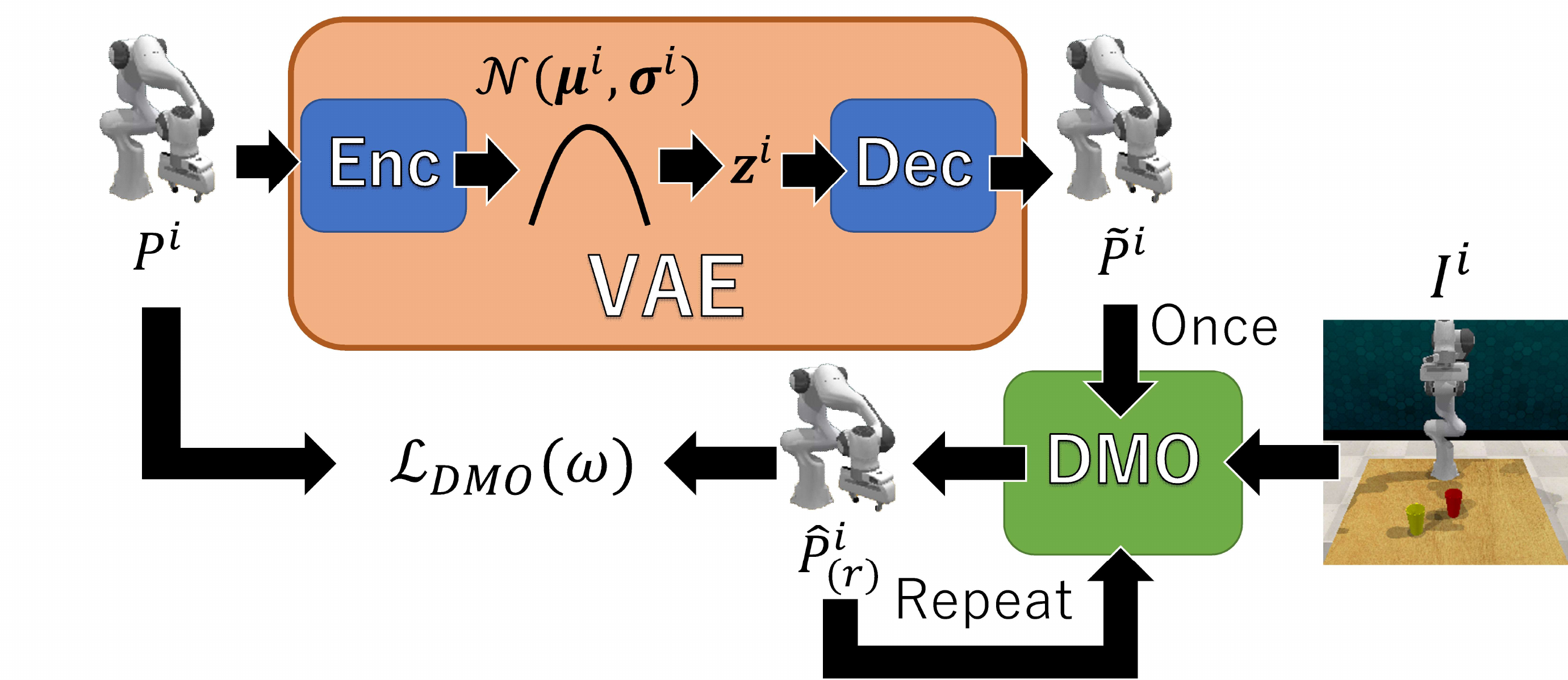}
    \caption{Training procedure of our Deep Motion Optimizer (DMO). VAE reconstructs the motion from hidden vector $\bm{z}^{i}$ which include small noise. DMO is optimized to refine this small difference to precisely predict the ground truth motion.}
    \label{fig:train DMO}
  \end{center}
\end{figure}

Different from general motion optimizers such as MCMC sampling~\cite{roberts1996exponential}, DMO alleviates the difficulty of hyper-parameter tuning by being trained in a data-driven manner as follows.
%
To train DMO, $(I^{i}, P^{i})$ in which the task can be accomplished is given as the ground-truth positive sample in the training dataset.
DMO is trained so that any $P$ close to $P^{i}$ is updated toward $P^{i}$.
In reality, however, $P$ is not randomly distributed but should be limited within realistic robot motions (e.g., within range of articulated motion).

In our method, a set of such realistic $P$ around $P^{i}$ is synthesized via the trained VAE, $V$, as done in the training of EBM.
This training procedure is illustrated in Fig.~\ref{fig:train DMO}.
$P^{i}$ is fed into $V$ to get the motion sampled from $V$.
This sampled motion (denoted by $\tilde{P}^{i}$) is close to its original motion $P^{i}$, as described in Sec.~\ref{section: train EBM}.

By substituting $(I^{i}, \tilde{P}^{i})$ to $(I, \hat{P}_{(r)})$ where $r=0$ in Eq.~(\ref{eq:dmo_r}), DMO gets the updated motion $\hat{P}^{i}_{(1)}$.
As with the inference stage, DMO expressed by Eq.~(\ref{eq:dmo_r}) is recurrently used to get $\{ \hat{P}^{i}_{(2)}, \cdots, \hat{P}^{i}_{(R+1)} \}$.

With $\{ \hat{P}^{i}_{(1)}, \cdots, \hat{P}^{i}_{(R+1)} \}$,
the parameters of DMO denoted as $\omega$ are optimized with the following loss in which the difference between $\hat{P}^{i}_{(r)}$ and its ideal motion $P^{i}$ is minimized:
\begin{equation}
    \mathcal{L}_{DMO}(\omega) = \frac{1}{N} \sum_{i=1}^{N} \sum_{r=1}^{R+1} || P^{i} - \hat{P}^{i}_{(r)}||_2,
\end{equation}
where $R$ denotes the number of recurrently-updated motions.
In $\mathcal{L}_{DMO}(\omega)$, motion data augmentation is achieved by training all $(R+1)$ motions (i.e., $\hat{P}^{i}_{(1)}, \cdots, \hat{P}^{i}_{(R+1)}$) instead of training only $\hat{P}^{i}_{(R+1)}$.


\section{Network Architectures of EBM and DMO}
\label{section: proposed model}

\begin{figure}[t]
  \begin{center}
    \includegraphics[width=1.0\columnwidth]{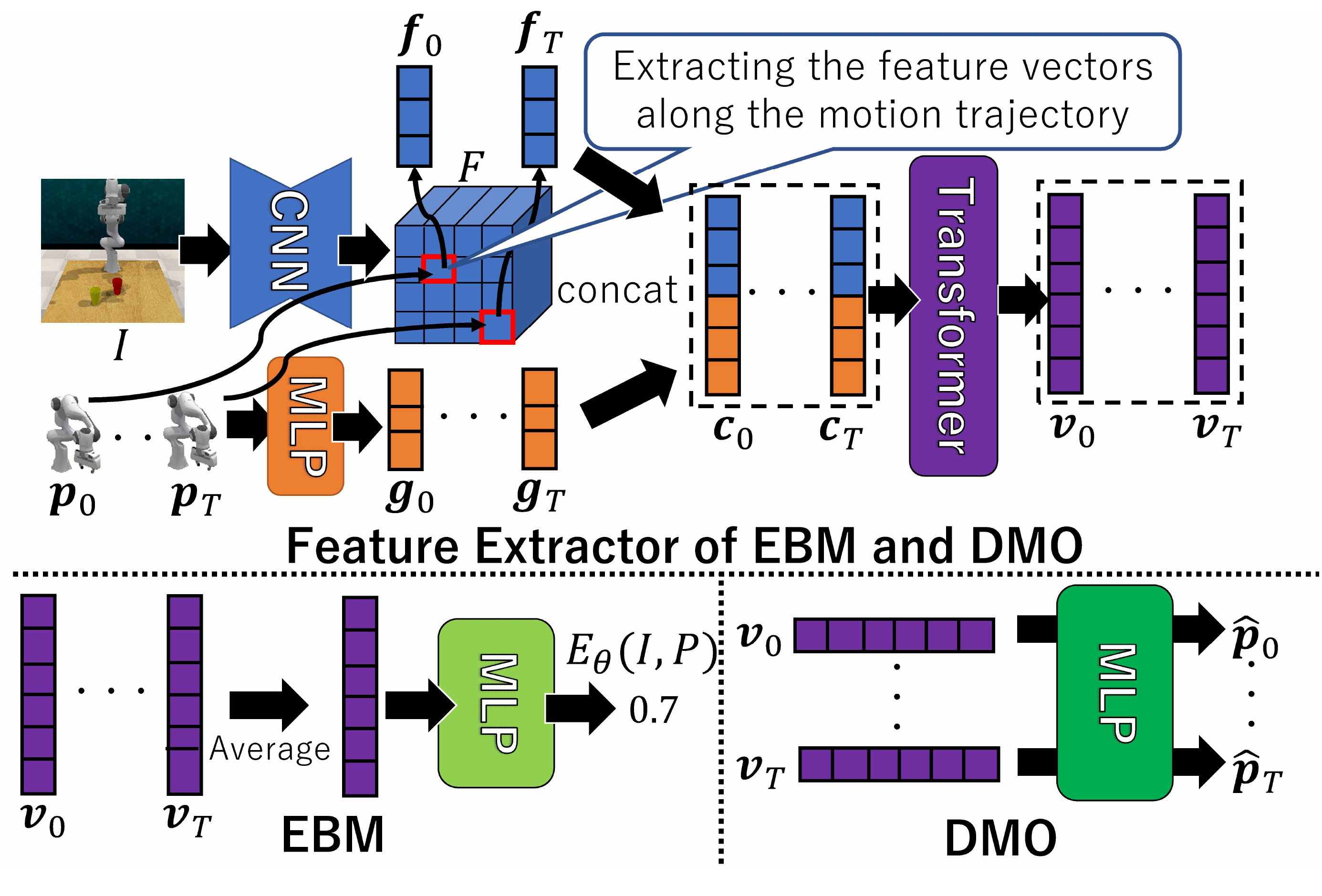}
    \vspace*{-6mm}
    \caption{The architectures of EBM and DMO. EBM and DMO have the same feature extractor but the parameters are not shared. CNN has the UNet-shaped architecture to create the feature map $F$. The pose features are extracted by MLP. To merge the image and pose features, the image features $\bm{f}_{k}$ are extracted from $F$ along the motion trajectory, and concatenated with the pose features $\bm{g}_{k}$. These features are fed into Transformer to provide the features $\bm{v}_{k}$ to EBM and DMO.}
    \label{fig:model architecture}
  \end{center}
\end{figure}

EBM and DMO have the same architecture for feature extraction (as shown in the upper part of Fig.~\ref{fig:model architecture}) but are trained independently, while their last layers are differently designed (as shown in the lower part of Fig.~\ref{fig:model architecture}).
The UNet-shaped image feature extractor~\cite{ronneberger2015u} using CNN~\cite{liu2022convnet} extracts the feature maps (denoted by $F$) from $I$, as indicated by blue in Fig.~\ref{fig:model architecture}.
The spatial dimensions of $I$ and $F$ are equal.
Each pose vector $\bm{p}_{t}$ where $t \in \{ 0, \cdots, T \}$ is independently fed into the MLP-based pose feature extractor, as indicated by orange in Fig.~\ref{fig:model architecture}.
Each extracted pose feature is denoted by $\bm{g}_{t}$.

$F$ and $\bm{g}_{t}$ are fed into Transformer for extracting informative features by merging $F$ and $\bm{g}_{t}$.
While all the features in $F$ can be theoretically fed into this Transformer, many redundant image features such as those of the background are less useful for motion prediction.
Such useless features make it difficult to properly train Transformer.
In our method, this problem is resolved by extracting only meaningful features from $F$.
Based on the idea that the most meaningful features for motion prediction are located along the motion trajectory, our method extracts the features in the temporal image coordinates of the robot hand, each of which is denoted by $\bm{u}_{t}$ at $t \in \{ 0, \cdots, T \}$. 
$\bm{u}_{t}$ is the 2D projection of $\bm{x}_{t}$, which is the 3D position of the robot hand.
Given the real coordinates $\bm{u}_{t}$, the feature in $\bm{u}_{t}$ (denoted by $\bm{f}_{t}$) is computed using bilinear interpolation in $F$.
$\bm{f}_{t}$ is concatenated with $\bm{g}_{t}$, and all the temporal concatenated features (each of which is denoted by $\bm{c}_{t}$) is fed into Transformer.
From $\{ \bm{c}_{0}, \cdots, \bm{c}_{T} \}$, Transformer extracts the temporal features $\{ \bm{v}_{0}, \cdots, \bm{v}_{T} \}$, as indicated by purple in Fig.~\ref{fig:model architecture}.

The advantages of the aforementioned concatenated features are summarized as follows:
\begin{enumerate}
\item By spatially aligning the image and pose coordinate systems, $\{ \bm{f}_{0}, \cdots, \bm{f}_{T} \}$ are extracted from the 2D positions consisting of each motion trajectory in $F$.
Since $\{ \bm{f}_{0}, \cdots, \bm{f}_{T} \}$ are temporal data, this encoding can be regarded as spatially-aligned temporal encoding.
\item Data of two different domains (i.e., image and motion) are integrated into the image feature domain.
\item The computational cost and training difficulty are reduced by decreasing the data size fed into Transformer.
\end{enumerate}

Given $\{ \bm{v}_{0}, \cdots, \bm{v}_{T} \}$, EBM and DMO are designed as follows:
\begin{itemize}
\item {\bf EBM:} EBM evaluates the likelihood of temporal data as the energy (denoted by $E_{\theta}(I, P)$ in Fig.~\ref{fig:model architecture}). This temporal evaluation is done by simultaneously feeding $\{ \bm{v}_{0}, \cdots, \bm{v}_{T} \}$ into the evaluation network, which is implemented by MLP in our method. For simultaneous input of $\{ \bm{v}_{0}, \cdots, \bm{v}_{T} \}$, their mean vector is fed into MLP for more efficient processing than other temporal processing such as recurrent networks.
\item {\bf DMO:} As with other temporal synthesis methods using Transformer, DMO separately accepts $\{ \bm{v}_{0}, \cdots, \bm{v}_{T} \}$ to independently acquire $\hat{\bm{p}}_{t}$ from $\bm{v}_{t}$.
\end{itemize}

Our code is available at github.com/Obat2343/DMOEBM.


\section{Experimental Results}

\subsection{Dataset}

\begin{figure*}[t]
  \begin{center}
    \includegraphics[width=\textwidth]{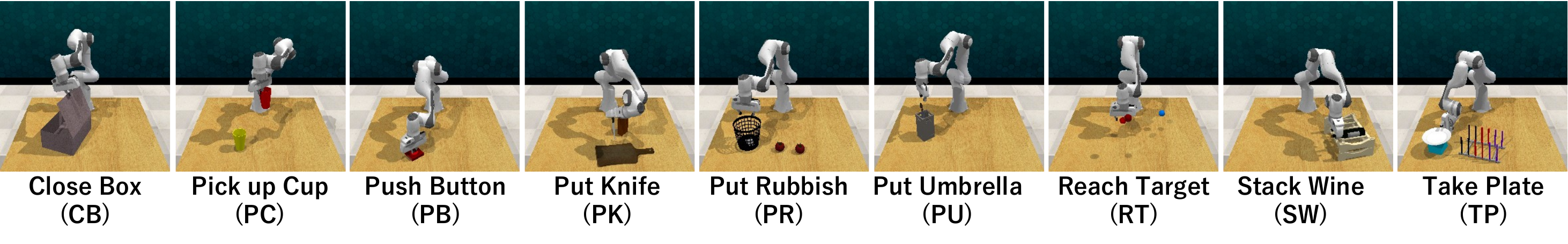}
    \vspace*{-6mm}
    \caption{List of the nine tasks. The position of the camera observing the initial image is fixed. Objects are randomly placed.}
    \label{fig:task list}
  \end{center}
  \vspace*{-4mm}
\end{figure*}

We evaluate our method using the RLBench dataset~\cite{james2019rlbench} in which a robot performs various tasks on the simulator, Pyrep~\cite{james2019pyrep}.
The nine tasks used in our experiments are shown in Fig.~\ref{fig:task list}.
For each task, 1,000 and 100 training and test sequences were generated in the simulator.
In all the sequences, the Franka Panda robot was used.
The initial state was captured as an image in each sequence.
Each image consists of RGB channels and a depth channel.
The image size is $256 \times 256$ pixels.
As VAE, ACTOR~\cite{petrovich2021action} was used.


\subsection{Motion Optimizers}

Our proposed motion optimizer (DMO) is evaluated with other motion optimizers so that each of the following optimizers is substituted for ``Motion Optimizer'' in Fig.~\ref{fig:inference}:

\noindent
{\bf (a) Ours-5:} DMO proposed in Sec.~\ref{section: train DMO}.
$R=5$ and $R=1$ in the training and prediction stages, respectively.

\noindent
{\bf (b) Ours-1:} DMO proposed in Sec.~\ref{section: train DMO}.
Both in the training and prediction stages, $R=1$.

\noindent
{\bf (c) Langevin~\cite{roberts1996exponential}:} Langevin MCMC is widely used with EBM for prediction.
For motion prediction, Langevin MCMC updates a motion using the gradient estimated by EBM and additive noise.
Two hyper-parameters, the step size and the number of iterations, are 0.001 and 100, respectively.

\noindent
{\bf (d) GD~\cite{wilson2003general}:} 
Different from Langevin MCMC, gradient descent (GD) optimization is not affected by noise.
As with (c) Langevin MCMC, the step size and the number of iterations, are 0.001 and 100, respectively.

\noindent
{\bf (e) VAEBM-L~\cite{xiao2021vaebm}:} Langevin MCMC is achieved in the embedding space trained by VAE.
The step size and the number of iterations, are 0.01 and 100, respectively.

\noindent
{\bf (f) VAEBM-S~\cite{xiao2021vaebm}:}
This is same with (e) except that the step size is 0.001.

\begin{table}[t]
\vspace*{1.5mm}
\caption{Task success rate of prediction methods.
\re{Red} and \cb{blue} values indicate the \re{best} and \cb{second-best}, respectively.}
\vspace*{-6mm}
\begin{center}
\scalebox{0.95}{
\begin{tabular}{l|c|c|c|c|c|c|c|c|c}
\hline
              & CB      & PC      & PB      & PK      & PR      & PU      & RT      & SW      & TP \\ \hline \hline
(a) Ours-5      & \re{97} & \re{89} & \cb{85} & \re{51} & \re{85} & \re{25} & \cb{32} & \cb{74} & \re{94} \\ \hline
(b) Ours-1      & \cb{95} & \re{89} & \re{87} & \cb{38} & \cb{77} & \cb{19} & 31      & \re{85} & \re{94} \\ \hline \hline
(c) Langevin  & 71      & 77      & 83      & 0       & 46      & 11      & 0       & 18      & 77      \\ \hline
(d) GD        & 54      & 78      & 61      & 18      & 40      & 5       & \re{36} & 48      & 81 \\ \hline
(e) VAEBM-L   & 64      & 88      & 81      & 27      & 35      & 10      & \cb{32} & 48      & 77 \\ \hline
(f) VAEBM-S   & 62      & 82      & 79      & 27      & 40      & 9       & \cb{32} & 50      & 74 \\ \hline
\end{tabular}
}
\end{center} \label{table:inference success rate}
\vspace*{-1.2mm}
\end{table}

\begin{figure}[t]
  \begin{center}
    \includegraphics[width=1.0\columnwidth]{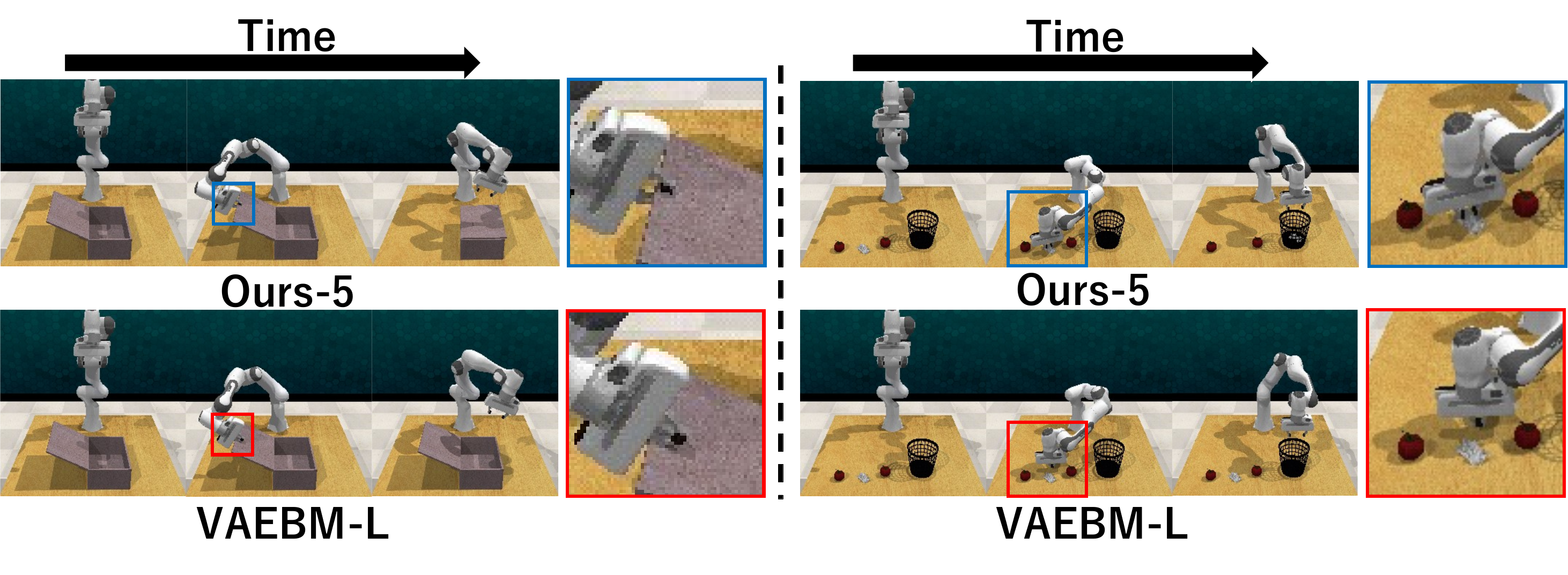}
    \vspace*{-6mm}
    \caption{Examples of predicted motions. Left: CB task. Right: PR task.}
    \label{fig:motion example}
  \end{center}
\end{figure}

The success rate of each optimizer is shown in Table~\ref{table:inference success rate}.
In all tasks except RT, our methods (a) and (b) outperform the others.
This verifies the superiority of the proposed combination of probabilistic and deterministic methods over the probabilistic methods (c), (d), (e), and (f).
The examples of motions predicted by (a) Ours-1 and (e) VAEBM-L, which is the best among the others, are shown in Fig.~\ref{fig:motion example}.
We can see that Ours-1 can finely rectify the predicted motions for accomplishing the tasks.
In comparison between (a) Ours-5 and (b) Ours-1, Ours-5 is a bit better.
This might be because of motion data augmentation by iterative motion updates.

More examples are shown in the supplementary video.






\subsection{Model Architectures of Feature Extractor}

\begin{figure}[t]
  \begin{center}
    \includegraphics[width=1.0\columnwidth]{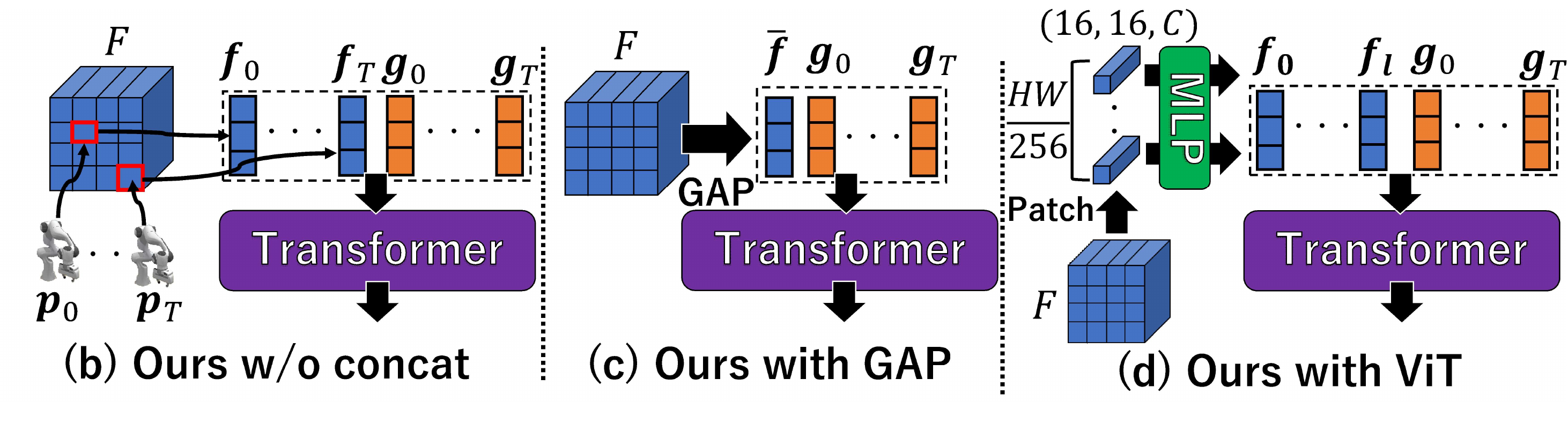}
    \vspace*{-6mm}
    \caption{Model architectures modified from our model.
    }
    \label{fig:model comparison}
  \end{center}
\end{figure}

The following four variants of the feature extractor in EBM and DMO are evaluated.

\noindent
{\bf (a) Ours:} The feature extractor with the proposed spatially-aligned temporal encoding (upper part of Fig.~\ref{fig:model architecture}).

\noindent
{\bf (b) Ours w/o concat:} Same with (a) except that $\{ \bm{g}_{0}, \cdots, \bm{g}_{T} \}$ are not concatenated with $\{ \bm{f}_{0}, \cdots, \bm{f}_{T} \}$.
All the features are directly fed into Transformer, as shown in Fig.~\ref{fig:model comparison} (b).

\noindent
{\bf (c) Ours with GAP:} $F$ is reduced to $\bar{\bm{f}}$ by Global Average Pooling~\cite{lin2013network} for dimensionality reduction. $\bar{\bm{f}}$ and $\{ \bm{g}_{0}, \cdots, \bm{g}_{T} \}$, which are not concatenated, are fed into Transformer, as shown in Fig.~\ref{fig:model comparison} (c).

\noindent
{\bf (d) Ours with ViT feature:} As with ViT~\cite{dosovitskiy2020vit}, $F$ is spatially divided to patch features denoted by $\{ \bm{f}_{0}, \cdots, \bm{f}_{l} \}$ where $l=256$ in our experiments.
The patch features are positionally encoded as done in generic Transformers~\cite{vaswani2017attention}, and then fed into Transformer with the pose vectors $\{ \bm{g}_{0}, \cdots, \bm{g}_{T} \}$ without concatenation, as shown in Fig.~\ref{fig:model comparison} (d).

\begin{table}[t]
\vspace*{1.5mm}
\caption{Task success rate of various architectures.
{\rm (a)} and {\rm (b)} use our proposed spatially-aligned temporal encoding.
}
\vspace*{-6mm}
\begin{center}
\scalebox{0.95}{
\begin{tabular}{l|c|c|c|c|c|c|c|c|c}
\hline
                & CB      & PC       & PB       & PK      & PR      & PU      & RT      & SW      & TP      \\ \hline \hline
(a) Ours        & \re{97} & \re{89}  & \cb{85}  & \cb{51} & \cb{85} & \re{25} & \re{32} & \cb{74} & \re{94} \\ \hline
(b) w/o concat  & \cb{96} & \cb{85}  & \re{94}  & \re{55} & \re{91} & \cb{13} & \cb{26} & \re{85} & \cb{93} \\ \hline \hline
(c) with GAP    & 5       & 3        & 0        & 4       & 1       & 0       & 3       & 5       & 0       \\ \hline
(d) with ViT    & 0       & 0        & 0        & 1       & 1       & 1       & 4       & 0       & 2       \\ \hline
\end{tabular}
}
\end{center} \label{table:model comparison accuracy}
\vspace*{-1.2mm}
\end{table}

Their task success rates are shown in Table~\ref{table:model comparison accuracy}.
Our spatially-aligned temporal coding (a) and (b) outperform (c) and (d), while  the effectiveness of the feature concatenation done in (a) is not validated.
%
%


\begin{table}[t]
\caption{The number of model parameters and computational cost.}
\begin{center}
\scalebox{0.95}{
\begin{tabular}{l|ll|ll}
\hline
\multirow{2}{*}{} & \multicolumn{2}{c|}{{\em All}}            & \multicolumn{2}{c}{{\em Reuse}}       \\ \cline{2-5} 
                  & \multicolumn{1}{l|}{Params (M)}     & Cost (GMac)        & \multicolumn{1}{l|}{Params} & Cost \\ \hline \hline
(a) Ours              & \multicolumn{1}{l|}{35.84}      & 23.62       & \multicolumn{1}{l|}{1.00}   & 0.11 \\ \hline
(b) Ours w/o concat   & \multicolumn{1}{l|}{35.68}      & 23.71       & \multicolumn{1}{l|}{0.84}   & 0.19 \\ \hline \hline
(c) Ours with GAP     & \multicolumn{1}{l|}{35.68}      & 23.61       & \multicolumn{1}{l|}{0.84}   & 0.10 \\ \hline
(d) Ours with ViT     & \multicolumn{1}{l|}{35.87}      & 24.97       & \multicolumn{1}{l|}{0.84}   & 0.38 \\ \hline
\end{tabular}
}
\end{center} \label{table:model comparison time}
\end{table}

Table~\ref{table:model comparison time} shows the cost (i.e., the number of parameters and the computational cost)
of the four variants.
Table~\ref{table:model comparison time} shows only the results of EBM because the dominant part is shared by EBM and DMO.
Furthermore, when multiple motions are evaluated and updated with the same image by EBM and DMO, respectively, we can reuse $F$ for reducing the cost.
The results in the cases where $F$ is reused and not reused are shown in {\em Reuse} and {\em All} columns, respectively.

The costs are almost same between (a), (b), (c), and (d) because the cost in CNN is larger than the costs in Transformer and MLP.
On the other hand, it can be confirmed that the costs in {\em Reuse} are much smaller than those in {\em All}.
This property is important
because the performance is gained as the number of input motions (i.e., $n$) increases, as demonstrated later in Table~\ref{table:inference parameter n}.
If the number of input motions becomes $n$-fold, the total cost in Reuse also becomes $n$-fold.



\subsection{EBM Training Methods}

The following two variants of the EBM training methods are evaluated, while their prediction method is fixed to be that described in Sec.~\ref{section: inference}.

\noindent
{\bf (a) Ours:} The training procedure shown in Fig.~\ref{fig:train EBM}.
For making a negative pair with $I^{i}$, either of $P^{j}$ where $j \neq i$ or $\tilde{P}^{j}$ synthesized by VAE is paired with $I^{i}$.

\noindent
{\bf (b) Ours with only VAE:} This is the same with (a) except that only $\tilde{P}^{j}$ synthesized by VAE is paired with $I^{i}$ for making negative pairs, as done in~\cite{xiao2021vaebm} and~\cite{gao2020flow}.

\begin{table}[t]
\caption{Task success rates of two variants of our prediction model.}
\vspace*{-6mm}
\begin{center}
\scalebox{0.95}{
\begin{tabular}{l|c|c|c|c|c|c|c|c|c}
\hline
                & CB      & PC    & PB       & PK      & PR      & PU      & RT      & SW      & TP  \\ \hline \hline
(a) Ours            & 95      & 86    & 73       & 56      & 84      & 13      & 29      & 79      & 74  \\ \hline
(b) Only VAE        & 0       & 3     & 5        & 3       & 2       & 0       & 0       & 0       & 0   \\ \hline
\end{tabular}
}
\end{center} \label{table:trainning method comparison}
\end{table}

The task success rates of these two variants are shown in Table~\ref{table:trainning method comparison}.
The scores in (b) are significantly worse.
We interpret this performance drop in what follows.

In (b), the loss is expressed by $\sum_{j \neq i} \nabla E_{\theta}(I^i,P^j)$ is removed from Eq.~(\ref{eq:EBM_loss}).
This loss can be minimized not only if any $(I, P)$ can be correctly discriminated between positive and negative data but also if $\tilde{P}$ synthesized by $V$ is discriminated from $P^{i}$ independently of $I^{i}$.
This latter case might be caused because the loss is decreased to almost zero, while the task success rate is low.
%
%
This problem can be resolved by synthesizing training data that cannot be discriminated from $P^{i}$.
While a possible way to synthesize such realistic training data is gradient-based adversarial training such as GAN~\cite{goodfellow2020generative}, it is unstable~\cite{gulrajani2017improved, miyato2018spectral}.
We also confirmed that EBM training methods using Langevin MCMC~\cite{roberts1996exponential} and VAEBM~\cite{xiao2021vaebm}, both of which exploit gradient information, are unstable.
Instead of synthesizing realistic data, on the other hand, real training data is used as negative samples, as expressed in Eq.~(\ref{eq:EBM_loss}).


\subsection{Parameters in DMO Inference}
\label{section: result change parameter}

\begin{table}[t]
\caption{Task success rates of our method with different $n$.}
\vspace*{-6mm}
\begin{center}
\scalebox{0.95}{
\begin{tabular}{l|c|c|c|c|c|c|c|c|c}
\hline
                & CB      & PC      & PB      & PK      & PR      & PU      & RT      & SW      & TP   \\ \hline \hline
$R$=1, $n$=1    & \re{99} & 83      & \cb{84} & 37      & 54      & 14      & \re{32} & 57      & 84   \\ \hline
$R$=1, $n$=8    & \cb{97} & \re{89} & \re{85} & \cb{51} & \cb{85} & \cb{25} & \re{32} & \cb{74} & \cb{94} \\ \hline
$R$=1, $n$=1000 & \cb{97} & \cb{88} & 68      & \re{71} & \re{86} & \re{26} & \re{32} & \re{83} & \re{95} \\ \hline
\end{tabular}
}
\end{center} \label{table:inference parameter n}
\vspace*{2mm}
\caption{Task success rates of our method with different $R$.}
\vspace*{-2mm}
\begin{center}
\scalebox{0.95}{
\begin{tabular}{l|c|c|c|c|c|c|c|c|c}
\hline
                & CB      & PC      & PB      & PK      & PR      & PU      & RT      & SW      & TP   \\ \hline \hline
$R$=1, $n$=8    & \re{97} & \re{89} & \re{85} & 51      & \re{85} & \re{25} & \re{32} & 74      & \re{94} \\ \hline
$R$=5, $n$=8    & \cb{95} & 86      & 73      & \cb{56} & \cb{84} & 13      & \cb{29} & \re{79} & 74   \\ \hline
$R$=10, $n$=8   & \cb{95} & \cb{88} & \cb{76} & \re{57} & 78      & \cb{16} & 7       & \re{79} & \cb{92} \\ \hline
\end{tabular}
}
\end{center} \label{table:inference parameter r}
\end{table}

In Table~\ref{table:inference parameter n} and ~\ref{table:inference parameter r}, the performance changes with the changes in $n$ and $R$ are shown, respectively.
%
By increasing $n$, the performance is gained in many tasks such as PC, PK, PR, PU, SW, TP, while $n$ has less impact in CB and RT.
Since $n$ is the number of motions evaluated in EBM and rectified in DMO, it is natural that the task success rate can be increased.
By increasing $R$, on the other hand, the performance change is fluctuating.
This might be caused because a limited number of training data might be insufficient, if the task is achievable by a variety of motions.
In such a case, since the training samples are sparsely distributed, overfitting is caused by increasing the number of iterative updates (i.e., $R$).


\section{Concluding Remarks}

This paper proposed a robot motion prediction method using a given image representing the initial situation of the environment including the robot.
Our method modifies EBM for evaluating the consistency between data of different domains (i.e., image and motion data).
In addition, a data-driven motion optimizer called DMO is proposed for improving the task achievability in a high-dimensional motion space.

Future work includes broader motion augmentation.
Motion samples augmented by VAE for training EBM and DMO are locally distributed.
However, broader motion samples are needed for predicting more complex motions.
For example, the diffusion model~\cite{ho2020denoising} is a prospective solution.



\bibliographystyle{IEEEtran}
\bibliography{icra_final}

\end{document}